\documentclass[10pt, a4paper]{article}
\usepackage{lrec}
\usepackage{graphicx}
\usepackage{tabularx}
\usepackage{soul}
\usepackage{xcolor}

\usepackage{epstopdf}
\usepackage[utf8]{inputenc}

\usepackage{hyperref}
\usepackage{xstring}

\newcommand{\code}[1]{{\texttt{#1}}}

\usepackage{float}
\newcommand{\af}[1]{#1}
\newcommand{\lu}[1]{#1}
\newcommand{\cm}[1]{#1}
\newcommand{\jh}[1]{#1}
\title{\af{Automatic Creation of Text Corpora for Low-Resource Languages from the Internet: The Case of Swiss German}}

\name{%
Lucy Linder$^{1}$, 
Michael Jungo$^{1}$,
Jean Hennebert$^{1}$,
Claudiu Musat$^{3}$,
Andreas Fischer$^{1,2}$}

\address{%
$^{1}$ iCoSys Institute, University of Applied Sciences and Arts Western Switzerland \\
$^{2}$ DIVA Group, University of Fribourg, Switzerland \\
$^{3}$ Data, Analytics and AI Lab, Swisscom AG, Switzerland \\ 
\{lucy.linder, michael.jungo, jean.hennebert, andreas.fischer\}@hefr.ch, claudiu.musat@swisscom.com 
} 

\abstract{
\af{This paper presents SwissCrawl, the largest Swiss German text corpus to date. Composed of more than half a million sentences, it was generated using a customized web scraping tool that could be applied to other low-resource languages as well. \jh{The approach demonstrates }\cm{how freely available web pages can be used to construct} comprehensive text corpora, which are of fundamental importance for natural language processing. In an experimental evaluation, we show that using the new corpus leads to significant improvements for the task of language modeling. 
\jh{To capture new content, our approach will run continuously} to keep increasing the corpus over time.}
\newline }

\begin{document}

\maketitleabstract

\section{Introduction}

Swiss German (``Schwyzerdütsch'' or ``Schwiizertüütsch'', abbreviated ``GSW'') is the name of a large continuum of dialects attached to the Germanic language tree spoken by more than 60\% of the Swiss population \af{\cite{ofs_langs}}. Used every day from colloquial conversations to business meetings, Swiss German in its written form has become more and more popular in recent years with the rise of blogs, \jh{messaging applications } 
and social media. 
\jh{However, the variability of the written form is rather large as orthography is more based on local pronunciations and emerging conventions than on a unique grammar.}

Even though Swiss German is widely spread in Switzerland, there are still few natural language processing (NLP) corpora, studies or tools available \cite{noah}. This lack of \af{resources} may be explained by the small pool of speakers (\af{less than one percent} of the world population), but also the many intrinsic difficulties of Swiss German, \af{including the lack of official writing rules, the high variability across different dialects, and the informal context in which texts are commonly written. Furthermore, there is no official top-level domain (TLD) for Swiss German on the Internet, which renders the automatic collection of Swiss German texts more difficult.}

To foster the development of NLP tools for Swiss German, 
we gathered the \af{largest} corpus of written Swiss German to date by crawling the web using a \af{customized} tool. \af{We highlight the difficulties for finding Swiss German on the web and demonstrate in an experimental evaluation how our text corpus can be used to significantly improve an important NLP task
: language modeling.}

\section{\af{Related Work}}

Few GSW corpora already exists. \af{Although they are very valuable for research on specific aspects of the Swiss German language,} they are either highly specialized \cite{sms-corpus} \cite{archimob} \cite{sb-ch}, \af{rather small} \cite{noah} (7,305 sentences), or do not offer full sentences \cite{crubadan}.

To our knowledge, the only comprehensive written Swiss German corpus to date comes from the Leipzig \af{corpora collection} initiative~\cite{leipzig} \af{offering corpora for more than 136 languages. The} Swiss German data \af{has} two sources: \af{the} Alemannic Wikipedia and web crawls on the \code{.ch} domain in 2016 and 2017, \af{leading to a total of} 175,399 unique sentences. While the Leipzig Web corpus for Swiss German is of considerable size, we believe this number \af{does not} reflect the actual amount of GSW available on the \af{Internet}. \lu{Furthermore, the enforced sentence structures do not represent the way Swiss German speakers write online}. 

In this paper, we thus aim at augmenting the Leipzig Web corpus by looking further than the \code{.ch} domain and by using a \af{suite of tools specifically designed for retrieving} Swiss German.


The idea of using the web as a vast source of linguistic data has been around for decades~\cite{Kilgarriff03introductionto} and many authors have already addressed its importance for low-resources languages \cite{Ghani2001MiningTW}. A common technique is to send queries made of mid-frequency $n$-grams to a search engine to gather bootstrap URLs, which initiate a \emph{crawl} using a breadth-first strategy in order to gather meaningful information\lu{,} such as documents or words \cite{Sharoff2006}, \cite{crubadan}.

Existing tools and studies, however, have requirements that are inadequate for the \af{case of} Swiss German. For example, GSW is not a language known to search engines \cite{Sharoff2006}, \af{does not} have specific TLDs \cite{Schafer2012}, and lacks good language identification models. \jh{Also, GSW documents are too rare to use bootstrapping techniques} \cite{Ghani2001MiningTW}.
\jh{Finally}, as GSW is scarce and mostly found in comments sections or as part of multilingual web pages (e.g. High German), we cannot afford to ``privilege precision over recall'' \cite{KilgarriffBaroni2006} by focusing on the main content of a page.

\jh{As a consequence,} our \af{method} is based on known techniques \jh{that are} adapted to deal with those peculiarities. Furthermore, it was designed \af{for having a human in the loop. Its} iterative nature makes it possible to refine each step of the \af{tool chain} as \lu{our knowledge of GSW improves}.

\section{\af{Proposed System}}



\begin{figure}[t]
\begin{center}
\includegraphics[width=0.8\columnwidth]{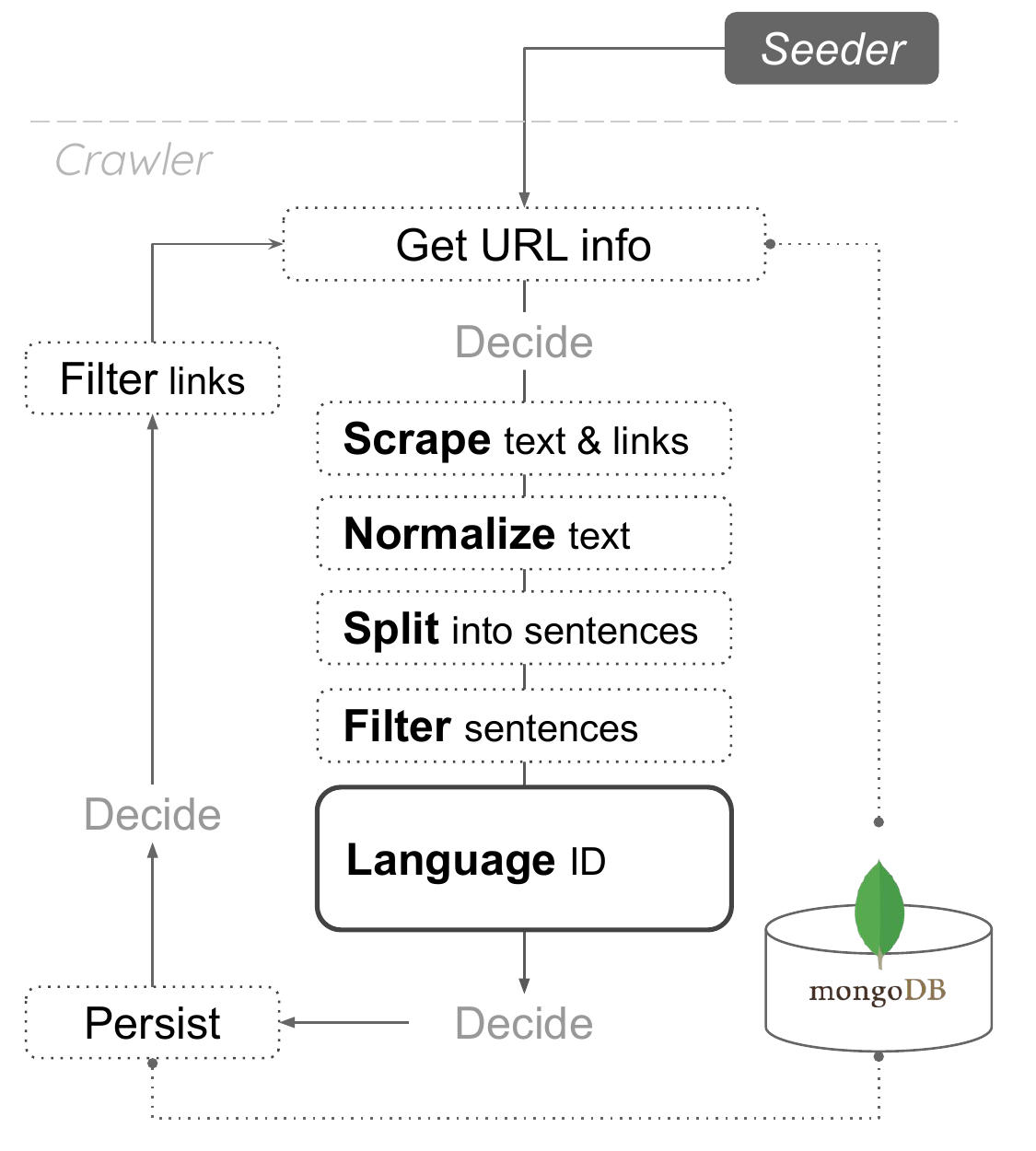}%
\caption{\af{Overview of one iteration of the proposed system.}}%
\label{fig:crawler-architecture}%
\end{center}
\end{figure}

\af{The two main components of our proposed system are shown in Figure~\ref{fig:crawler-architecture}:} a \emph{seeder} that gathers potentially interesting URLs using a Search Engine and a \emph{crawler} that extracts GSW from web pages, linked together by a MongoDB database. \af{The system} is implemented in Python~3, with the full code available on GitHub\footnote{\url{https://github.com/derlin/swisstext-lrec}}. 
Due to the exploratory nature of the task, the tool chain is executed in an iterative manner, allowing us to control and potentially improve the process punctually.

\subsection{The Seeder}

Query generation has already been extensively studied \cite{Baroni2006}, \cite{Sharoff2006}. In the case of Swiss German, we tested three different approaches: (a)~most frequent trigrams, (b)~selection of 2 to 7 random words weighted by their frequency distribution and (c)~human-generated queries.

\af{When comparing the corpora generated by 100 seeds of each type, we did not observe significant differences in terms of quantity or quality for the three seeding strategies.} 
\jh{On a positive side, $50\%$ of the sentences were different from one seed strategy to the other, suggesting for an approach where strategies are mixed.}
\jh{However, we also observed that} (a)~tends to yield more similar queries over time and (c)~is too time-consuming for practical use. 

\jh{Considering these observations,} we privileged the following approach:
\begin{enumerate} 
    \item Start with a list of sentences, either from a bootstrap dataset or from sentences from previous crawls using one single sentence per unique URL;
    \item Compute the frequency over the vocabulary, \jh{normalizing words to lower case} \lu{and discarding those having non-alphabetic characters;}
    \item Filter out words appearing only once or present in German or English vocabularies\footnote{Free German Dictionary, \url{https://sourceforge.net/projects/germandict/} 1.9M words and US word list from GNU Aspell \url{http://aspell.net/}, 40K words.};
    \item Generate query seeds by sampling 3 words with a probability following their frequency distribution; 
    \item Exclude seeds with more than two single-letter words or having a GSW probability below 95\% (see the language identification (LID) component, Section~\ref{lid}).
\end{enumerate}

Initial sentences come from the Leipzig \af{web corpus} 2017, filtered \af{by means of the LID described in Section~\ref{lid}}

Each seed is submitted to startpage.com, a Google Search proxy augmented with privacy features. To ensure GSW is not auto-corrected to High German, each word is first surrounded by \lu{double} quotes. The first 20 new URLs, i.e. URLs that were never seen before, are saved for further crawling. 

\subsection{The Crawler}\label{crawler-components}

\begin{table*}[ht]
    \centering
    \begin{tabular}{c|rrrr|rrr|r}
    \multicolumn{1}{c}{} &
    \multicolumn{4}{c}{\textit{Seeding}} &
    \multicolumn{3}{c}{\textit{Crawling}} &
    \multicolumn{1}{c}{\textit{}}\\ 
    \hline
    Iter. & Seeds & Found & Good & \%Good & Sentences  & Domains & URLs & Runtime \\ \hline
    0     & 100   & 837   & 577  & 68.94  & 89350 & 529  & 4810 & 16h15 \\ 
    1     & 100   & 1062  & 662  & 62.34  & 48483 & 552  & 4382 & 16h52 \\ 
    2   & 100   & 732   & 423  & 57.79  & 20662 & 423  & 2193 & 40h52 \\ \hline
    \end{tabular}%
    \caption{Number of URLs found vs actually pertinent during seeding and number of unique new sentences, domains, and URLs discovered during crawl starting from a blank database and launching the system three times.}%
    \label{tab:lrec-experiment}%
\end{table*}

The crawler starts with a list of URLs and metadata taken either from a file or from the MongoDB instance, and are added to a task queue with a depth of 0. As \af{illustrated} in Figure~\ref{fig:crawler-architecture}, 
\lu{each task consists of a series of steps that will download the page content, extract well-formed GSW sentences and add links found on the page to the task queue}.
At different stages of \lu{this} pipeline, a \emph{decider} can intervene in order to stop the processing early. A crawl may also be limited to a given depth, usually set to 3.

\paragraph*{Scrape} The raw HTML content is fetched and converted to UTF-8 using a mixture of \code{requests} and \code{BeautifulSoup}. Boilerplate removal such as navigation and tables uses \code{jusText}~\cite{pomikalek2011justext}, but ignores stop words filtering as such a list is not available for GSW. The output is a UTF-8 text.

\paragraph*{Normalize} This stage tries to fix remaining encoding issues using \code{ftfy}~\cite{speer-2019-ftfy} and to \jh{remove} \lu{u}nicode emojis. Another important task is to normalize the \lu{u}nicode code points used for accents, spaces, dashes, quotes etc., and strip any invisible characters. 
To further improve the usability of the corpus \jh{and to simplify tokenization}, we also try to enforce one single convention for spaces around quotes and colons, e.g. colons \emph{after} closing quote, no space inside quotes.

\paragraph*{Split} To split text into sentences, we implemented Moses' \code{split-sentences.perl}\footnote{\url{https://github.com/moses-smt/mosesdecoder}.} in Python and changed it in three main ways: existing newlines are preserved, colons and semi-colons are considered segmentation hints and sentences are not \af{required} to start with an uppercase. The latter is especially important as GSW is mostly found in comments where people tend to write fast and without proper casing/punctuation. The list of non-breaking prefixes used is a concatenation of the English and German prefixes found in Moses with few additions.

\paragraph*{Filter} Non- or bad- sentences are identified based on a list of $20+$ rules that normal sentences should obey. Most rules are specified in the form of regular expression patterns and boundaries of acceptable occurrences, few compare the ratio of occurrence between two patterns. Examples of such rules in natural language are: 
``no more than one hashtag'', 
``no word with more than 30 characters'',
``the ratio capitalized/lowercase words is below 1.5''.

\paragraph*{Language ID} Using the language identification model described in \af{Section}~\ref{lid}, sentences with a GSW probability of less than 92\% are discarded. This threshold is low \af{on purpose} in order to favor recall over precision.

\paragraph*{Link filter} This component is used to exclude or transform outgoing links found in a page based on duplicates, URL composition, but also specific rules for big social media sites or known blogs. Examples are the exclusion of unrelated national TLDs (\code{.af}, \code{.nl}, \ldots) and known media extensions (\code{.pdf}, \code{.jpeg}, etc.), the stripping of \af{session IDs} in URL parameters, \af{and} the homogenization of subdomains for sites such as Twitter. \af{Note that} filtering is based \af{only} on the URL \af{and therefore does not} handle redirects or URLs pointing to the same page. This \af{leads to} extra work during the crawling, but keeps the whole system simple.

\paragraph*{Decide} A decider has three main decisions to take. First, based on the metadata associated with an URL, should it be \af{visited?} In practice, we visit only new URLs, but the tool is designed \af{in a way such that} recrawls would be triggered depending on how much new content is found over time. 
The second decision arises at the end of the processing, where the page can be either saved or blacklisted. To favor recall, we currently keep any URL with at least one GSW sentence. Finally, the decider can choose to visit the outgoing links or not. After some trials, we found that following links from pages with more than two new GSW sentences \af{is a reasonable choice}\lu{, as p}\af{ages with less sentences} are often quotes or false positives.

\paragraph*{Duplicates} During the crawl, the uniqueness of sentences and URLs considers only exact matches. However, when exporting the results, near-duplicate sentences are removed by first stripping any non-letter (including spaces) and making a lowercase comparison. We tried other near-duplicate approaches, but found that \af{they} also discarded meaningful writing variations. 

\subsection{\af{Language Identification}}\label{lid}

\af{Language identification} (LID) is a central component of the pipeline, as it has a strong influence on the final result. In addition, readily available tools are not performing at a satisfying level. For these reasons we created a tailor-made LID system for this situation.

\af{LID} has been extensively studied over the past \af{decades}~\cite{jauhiainen2019automatic} and has achieved impressive results on long monolingual documents in major languages such as English.
However, the task becomes more challenging when the pool of training data is small \af{and of high variability, and when} the unit of identification is only a sentence.

Free pretrained LIDs supporting GSW such as FastText~\cite{joulin2016fasttext} are trained on the Alemannic Wikipedia, which encompasses not only GSW, but also German dialects such as Badisch, \af{Elsässisch}, Schwäbisch and \af{Vorarlbergisch}. This makes the precision of the model \af{insufficient for our purposes}.

\af{The dataset used \jh{to build our} Swiss German LID is based on the Leipzig text corpora~\cite{leipzig}, mostly focusing on the texts gathered from the Internet. In preliminary experiments, we have chosen eight language classes shown in Table~\ref{tab:lid-classes}, which give precedence to languages closely related to Swiss German in their structure. \jh{In this Table, \code{GSW\_LIKE} refers to a combination of} dialects that are similar to Swiss German but for which we did not have sufficient resources to model classes on their own.
}

\begin{table}[t]
    \begin{tabularx}{\columnwidth}{|>{\small \texttt}r|X|}
          \hline
          \textbf{Label} & \textbf{Languages} \\ \hline
          AFR & Afrikaans  \\ 
          DEU & German \\ 
          ENG & English \\  
          GSW & Swiss German \\  
          GSW\_LIKE & Bavarian, Kölsch, Limburgan, Low German, Northern Frisian, Palatine German\\  
          LTZ & Luxembourgian \\ 
          NLD & Dutch \\  
          OTHER & Catalan, Croatian, Danish, Esperanto, Estonian, Finnish, French, Irish, Galician, Icelandic, Italian, Javanese, Konkani, Papiamento, Portuguese, Romanian, Slovenian, Spanish, Swahili, Swedish \\ \hline
    \end{tabularx}%
    \caption{Composition of the \af{eight} classes used for \af{language identification}.}%
    \label{tab:lid-classes}%
\end{table}

\af{A total of 535,000 sentences are considered for LID with an equal distribution over the eight classes. The 66,684 GSW sentences originate from the Leipzig web corpus 2017 and have been refined during preliminary experiments to exclude obvious non-GSW contents. We use 75\% of the data for training, 10\% for optimizing system parameters, and 15\% for testing the final performance.}

\af{Using a pretrained German BERT model \cite{devlin2018bert}\footnote{\code{bert-base-german-cased} model from \url{https://github.com/huggingface/transformers}
\cite{wolf2019huggingface}
.} and fine-tuning it on our corpus, we obtain a high LID accuracy of 99.58\%. GSW is most confused with German (0.04\%) and \code{GSW\_LIKE} (0.04\%). We have also validated the LID system on SMS sentences~\cite{sms-corpus}, where it proves robust for sentences as short as five words.}

\section{State of the Swiss German Web}

Table~\ref{tab:lrec-experiment} shows the results of running the system three times using 100 different seeds on a virtual machine with 5 CPU cores and no GPUs. As expected, the first iteration yields the most new sentences. \af{Unfortunately}, the number of newly discovered hosts and sentences decreases exponentially as the system runs, dropping to 20K sentences on the third iteration. This \af{result emphasizes the fact} that the amount of GSW on the web 
is \af{very} limited. 

The third iteration took also \af{significantly} longer, which highlights 
the difficulties of crawling the web. In this \jh{iteration,} 
some URLs had as much as 12 \af{thousand} outgoing links that we had to visit before discarding. Another problem arises on web sites \jh{where query parameters are used in URLs to encode cookie information and on which duplicate hypotheses cannot be solved unless visiting the links.}

On each new \af{search engine} query, we go further down the list of results as the top ones may already be known. As such, the percentage of pertinent URLs retrieved (\% good, \lu{see decider description in Section~\ref{crawler-components}}) slowly decreases at each iteration. It is however still above 55\% of the retrieved URLs on the third run, indicating that further runs may still be beneficial.

\section{The SwissCrawl Text Corpus}

Using \af{the proposed system,} we were able to gather more than \af{half a million} unique GSW sentences from around the web. The crawling took place between September and November 2019. The \af{corpus is available for download}\footnote{\af{\url{https://icosys.ch/swisscrawl}}} in \lu{the} \af{form of} a CSV file with four columns: 
\code{text}, \code{url}, \code{crawl\_proba}, \code{date},
with \code{crawl\_proba} being the GSW probability returned by the LID \af{system} (see \af{Section}~\ref{lid}).

\subsection{Contents}\label{quantitative-analysis}

\begin{table}[ht]
\begin{tabular}{lrrc}
\hline
Domain     & \#URLs & \#Sentences & \%~    \\ \hline
www.fcbforum.ch   & 6,169  & 54,954  & 9.77  \\ 
www.wikiwand.com  & 2,404  & 36,432  & 6.48  \\ 
www.heiraten.ch   & 1,238  & 36,129  & 6.42  \\ 
forum.zscfans.ch  & 3,844  & 31,125  & 5.53  \\ 
www.celica-t23.ch & 3,007  & 26,446  & 4.70  \\ 
www.fczforum.ch   & 7,275  & 23,967  & 4.26  \\ 
www.facebook.com  & 3,280  & 17,601  & 3.13  \\ 
dict.leo.org      & 255   & 16,109  & 2.86  \\ 
twitter.com       & 2,720  & 9,061   & 1.61  \\ 
swizzlink.ch      & 1,000  & 7,465   & 1.33  \\ 
&&& \\
\textit{other}    & 31,329 & 303,235 & 53.91 \\
\hline
\end{tabular}%
\caption{The top ten domains found in the corpus.}%
\label{tab:top-hosts}%
\end{table}

\begin{figure}[ht]
    \centering
    \includegraphics[width=.8\columnwidth]{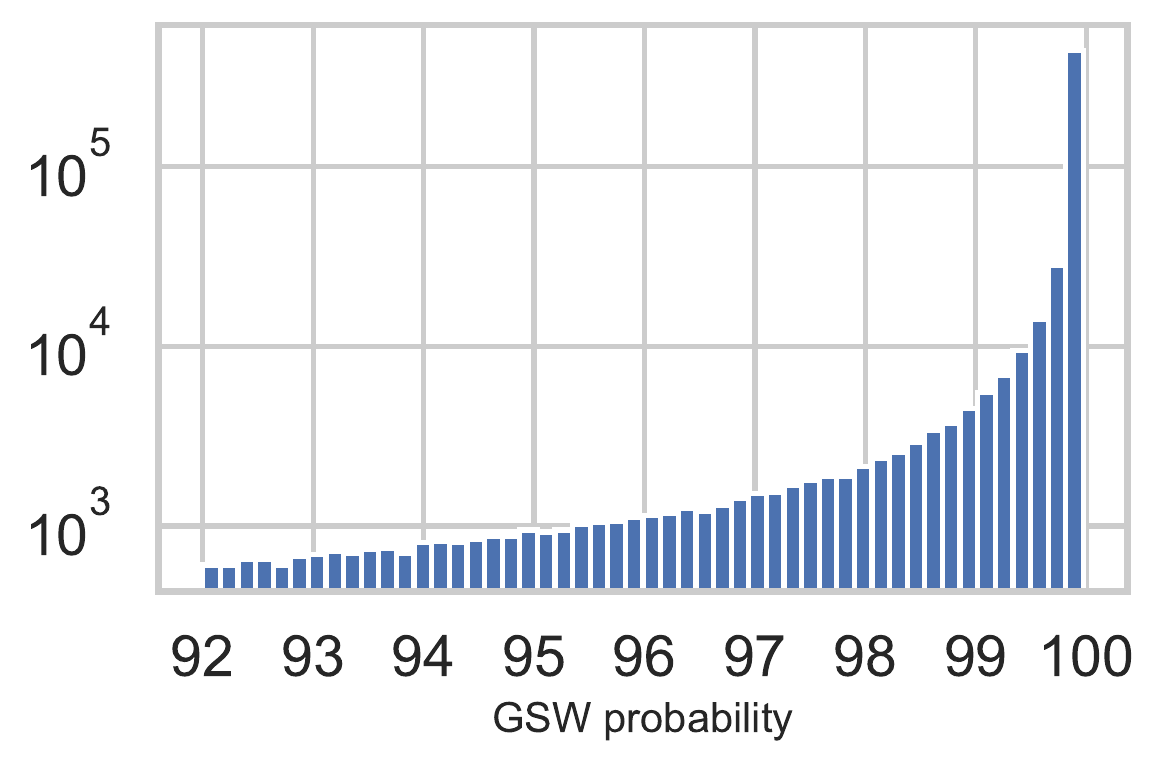}%
    \caption{Distribution of \code{crawl\_proba} {\footnotesize (logarithmic axis)}.}%
    \label{fig:crawl-proba}%
\end{figure}
\begin{figure}[ht]
    \centering
    \includegraphics[width=.8\columnwidth]{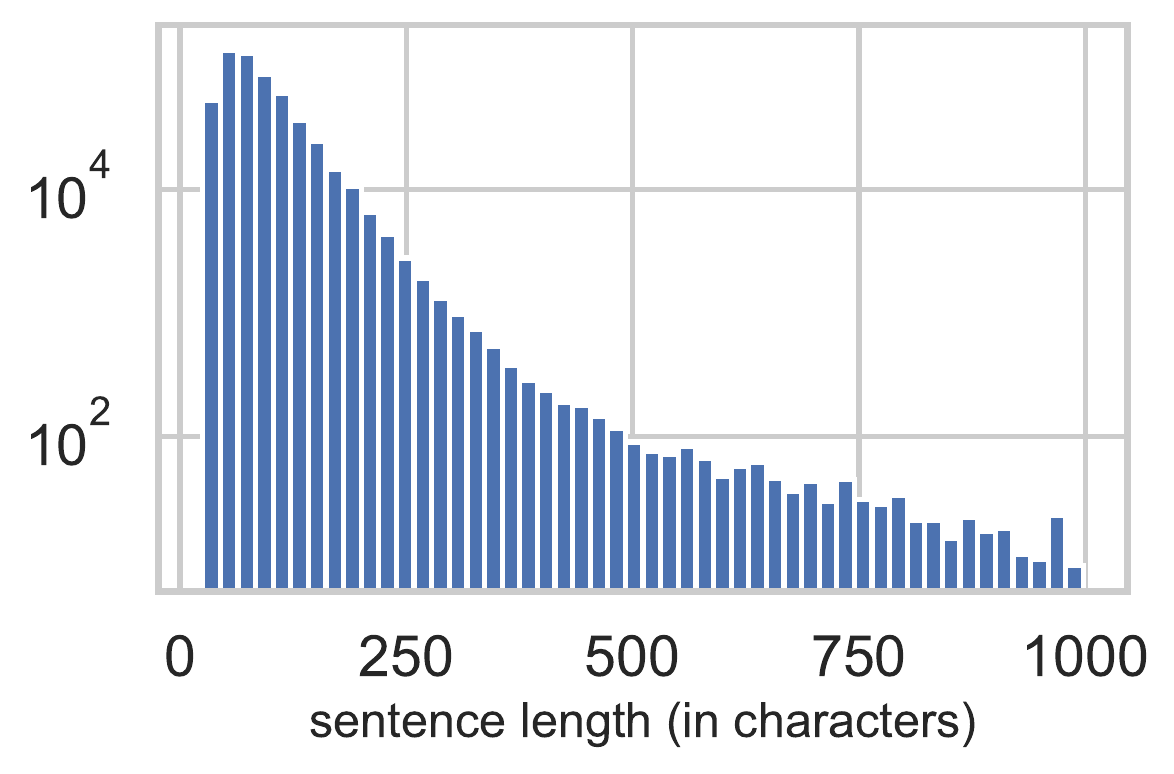}%
    \caption{Distribution of \code{text} length {\footnotesize (logarithmic axis)}.}%
    \label{fig:sentence-length}%
\end{figure}

\begin{table}[ht]
\centering
\begin{tabular}{rrr}
\hline
Proba. & \#Sentences & \%    \\ 
\hline
99+ & 503,526 & 89.51 \\ 
98+ & 19,769  & 3.51  \\ 
97+ & 11,064  & 1.97  \\ 
96+ & 7,940   & 1.41  \\ 
95+ & 6,367   & 1.13  \\ 
94+ & 5,344   & 0.95  \\ 
93+ & 4,541   & 0.81  \\ 
92+ & 3,973   & 0.71  \\ 
\hline
\end{tabular}%
\caption{Number of sentences for each 1\% probability bin.}%
\label{tab:crawl-proba}%
\end{table}

The corpus is composed of 562,524 sentences from 62K URLs among 3,472 domains. The top ten domains (see Table~\ref{tab:top-hosts}) are forums and social media sites\jh{. They} account for 46\% of the whole corpus.

In general, we consider a GSW probability of $\ge{99}\%$, to be \af{indeed Swiss German with high confidence}. This represents more than 89\% of the corpus (500K) (see Figure~\ref{fig:crawl-proba}). The sentence length varies between 25 and 998 characters with a mean of $92\pm55$ and a median of 77 (see Figure~\ref{fig:sentence-length}), while the number of words lies between 4 and 222, with a mean of $16\pm10$ and a median of 14. This highlights a common pattern in Swiss German writings: used mostly in informal contexts, sentences tend to be short and to include many symbols, \jh{such as emojis or repetitive punctuation.}

Very long sentences are usually lyrics that lack proper punctuation and thus \af{could not} be segmented properly. We however decided to keep them in the final corpus, as they could be useful in specific tasks and are easy to filter out otherwise.

Besides the normalization described in \ref{crawler-components}, no cleaning \jh{nor} post-processing is applied \af{to} the sentences. This is a deliberate choice to avoid losing any information that could be pertinent for a given task or for further selection. As a result, the mean letter density is 80\% and only 61\% of sentences both start with an uppercase letter and end with a common punctuation mark (\code{.!?}).

Finally, \af{although} we performed no human validation \af{\textit{per se}}, \af{we} actively monitored the crawling process to spot problematic domains early. This allowed to blacklist some domains entirely, for example those serving embedded PDFs (impossible to parse properly) or written in very close German dialects.

\subsection{Discussion}

\begin{table*}
    \centering
    \begin{tabularx}{\textwidth}{r|Xlr}
\hline 
\multicolumn{1}{c}{} & 
\multicolumn{1}{l}{Text} & Domain & Proba. \\ \hline
1 & E chlini Hommage a d Griächä, ihri kreativi Schprach und ihri relativ schrägä aber umso luschtigärä Brüch. & gurk.ch & 99.96 \\ 
2 & aso i würd nech no bis ändi nöchscht wuche chrank schribe. & twitter.com	& 99.96 \\ 
3 & heheheh aber nunuu das pic isch geil...:-) *hützobeeeeeeee* :-D Sa 2.9.06, 10: & www.festzeit.ch & 99.96\\ 
4 & Super Mario Odyssey \#14 - Rat wer zrugg isch... & www.youtube.com & 99.40 \\ 
5 & 14. Um(ge)kehrt ist au(ch) g'fahren. - Auerbach, Dorfgesch., III, 250; & www.zeno.org & 98.91\\  
6 & "Jungfrau Zeitung - Töffli-Revival über drei Pässe", "rh": & www.google.ch & 96.18\\\hline
    \end{tabularx}%
    \caption{Sample texts; 1-2 are of good quality, 3-4 contain \af{many special characters}, \af{5-6} are false \af{positives} (High German).}%
    \label{tab:qualitative-examples}%
\end{table*}

Table~\ref{tab:qualitative-examples} shows some hand-picked examples. As most of our sources are social medias and forums, the writing style is \lu{often} 
colloquial, interspersed with emojis and \af{slang}. This perfectly reflects the use of GSW in real life, where speakers switch to High German in formal conversations.

In general, the quality of sentences is good, with few false \af{positives} mostly in High German or German dialects, rarer still in Dutch or Luxembourgian. The presence of specific structures in the sentences are often the cause of such mistakes, as they yield strong GSW cues. For example:
\begin{itemize}
    \item High German with spelling mistakes or broken words;
    \item GSW named entities (\af{``Ueli Aeschbacher''}, ``Züri'');
    \item The presence of many umlauts and/or short words;
    \item The repetition of letters, also used to convey emotions.
\end{itemize}

The quality of the corpus highly depends on the text extraction step, which itself depends on the HTML structure of the pages. As there are no enforced standards and each website has its own needs, it is impossible \lu{to handle all edge cases}. 
For example, some sites use hidden \code{<span>} elements to hold information, which become part of the extracted sentences. This is true for watson.ch and was dealt \af{with using a specific rule}, but there are still instances we \af{did not} detect. 

Splitting text into sentences is not a trivial task. \jh{Typical} segmentation mistakes come from the use of ASCII emojis as punctuation marks (\af{see text sample 3} in Table~\ref{tab:qualitative-examples}), which are very common in forums. They are hard to detect 
\jh{due to the variability of each individual style.} 

We defined duplicates as having the exact same letters. As such, some sentences may differ by one umlaut and some may be the truncation of others (e.g. excerpts with ellipsis). Finally, the corpus also contains poems and lyrics. Sometimes repetitive and especially hard to segment, they are still an important source of Swiss German online. In any case, they may be filtered out using cues in the sentence length and the URLs.   

\section{Swiss German Language Modeling}

To demonstrate the effectiveness of the \lu{SwissCrawl corpus}, \af{we conducted a series of experiments for the NLP task of language modeling}. The whole code is publicly available on GitHub\footnote{\url{https://github.com/jungomi/swiss-language-model}}.

Using the GPT-2~\cite{radford2019language} model in its base configuration
(12 layers, 786 hidden states, 12 heads, 117M parameters), we trained three models using different training data:
\begin{enumerate}
    \item \textit{Leipzig} unique sentences from the Leipzig GSW web;
    \item \textit{SwissCrawl} sentences 
    \jh{with a GSW probability} $\ge{99}\%$ (see \af{Section}~\ref{quantitative-analysis});
    \item \textit{Both} the union of 1) and 2).
\end{enumerate}

For each model, the vocabulary is generated using Byte Pair Encoding (BPE)~\cite{sennrich-2015-bpe} applied on the training set.
BPE is a tokenization based on the most frequently occurring subword units, with the advantage of overcoming the out-of-vocabulary problem while still capturing meaningful information.
The \af{independent} test sets are composed of 20K samples from each source. 

Table~\ref{tab:lm-experiment} shows the perplexity of the models on each of the test sets. As expected, each model performs better on the test set they have been trained on. When applied to a different test set, both see an increase in perplexity. However, the Leipzig model seems to have more trouble generalizing: its perplexity nearly doubles on the SwissCrawl test set and raises by twenty on the combined test set.

The best results are achieved by combining both corpora: while the perplexity on our corpus only marginally improves (from $49.5$ to $45.9$), the perplexity on the Leipzig corpus improves significantly (from $47.6$ to $30.5$, a \af{36\% relative improvement}).

\begin{table}
    \begin{tabular}{cc|ccc}
      \multicolumn{2}{c}{\textit{Training}} &
      \multicolumn{3}{c}{\textit{Test Sets}} \\ \hline
      Dataset       & Size    & Leipzig  & \textit{SwissCrawl} &  Both \\ \hline
      Leipzig       & 180,000 & 47.6          & 92.6          & 67.6 \\
      \emph{SwissCrawl} & 483,526 & 63.9          & 49.5          & 56.2 \\
      Both & 663,526 & \textbf{30.5} & \textbf{45.9} & \textbf{38.0} \\ \hline
    \end{tabular}%
    \caption{Perplexity of language models trained on Leipzig, SwissCrawl (our corpus) and both.}%
    \label{tab:lm-experiment}%

\end{table}

\section{Conclusion}

\af{In this paper, we} presented the tools developed to gather the most comprehensive collection of written Swiss German to our knowledge. \af{It represents Swiss German in the way it is actually used in informal contexts, both with respect to the form (punctuation, capitalization, \ldots) and the content (slang, elliptic sentences, \ldots). We have demonstrated how this new resource can significantly improve Swiss German language modeling. We expect that other NLP tasks, such as LID and eventually machine translation, will also be able to profit from this new resource in the future.}

Our experiments support the reasoning that Swiss German is still scarce and very hard to find online. Still, the \af{Internet} is in constant evolution and we \af{aim} to keep increasing the corpus size by \af{rerunning the tool chain} \lu{at} regular intervals. Another \af{line of future development is the customization of the tools for} big social media platforms such as Facebook \af{and} Twitter, where most of the content is only accessible through specific APIs.

\bibliographystyle{lrec}
\bibliography{bibliography}

\end{document}